# SUPERVISED FEATURE SELECTION FOR DIAGNOSIS OF CORONARY ARTERY DISEASE BASED ON GENETIC ALGORITHM

Sidahmed Mokeddem[1], Baghdad Atmani[2] and Mostéfa Mokaddem[3]

Equipe de recherche SIF « Simulation, Intégration et Fouille de données », Laboratoire d'Informatique d'Oran, LIO, Oran University, Algeria

[1]`sid_wise@hotmail.com`
[2]`atmani.baghdad@gmail.com`
[3]`mmemus@gmail.com`

## ABSTRACT

*Feature Selection (FS) has become the focus of much research on decision support systems areas for which datasets with tremendous number of variables are analyzed. In this paper we present a new method for the diagnosis of Coronary Artery Diseases (CAD) founded on Genetic Algorithm (GA) wrapped Bayes Naïve (BN) based FS.*

*Basically, CAD dataset contains two classes defined with 13 features. In GA–BN algorithm, GA generates in each iteration a subset of attributes that will be evaluated using the BN in the second step of the selection procedure. The final set of attribute contains the most relevant feature model that increases the accuracy. The algorithm in this case produces 85.50% classification accuracy in the diagnosis of CAD. Thus, the asset of the Algorithm is then compared with the use of Support Vector Machine (SVM), Multi-Layer Perceptron (MLP) and C4.5 decision tree Algorithm. The result of classification accuracy for those algorithms are respectively 83.5%, 83.16% and 80.85%. Consequently, the GA wrapped BN Algorithm is correspondingly compared with other FS algorithms. The Obtained results have shown very promising outcomes for the diagnosis of CAD.*

## 1. INTRODUCTION

Some studies have proven that CAD is the leading cause of death in countries under development such as Algeria. It is estimated that 45% of deaths in Algeria are due to cardiovascular disease [7]. The risk of cardiovascular disease is linked by various factors such as environmental, psychological, genetic, demographic variables and health services. Many of these diseases require surgical treatment, most of these surgical interventions are expensive, and normal population cannot have such interventions. Moreover, the diagnosis of CAD is challenging, particularly when there is no symptom. Much information from patients is needed. Hence, the necessity of prevention systems and to predict risk factors for these diseases in order to take preventive measures is primordial.

With the increasing complexity in recent years, a large amount of information in the medical field is stored in electronic form such as the electronic patient record. These data are stored and used primarily for the management and analysis of the patient population. They are frequently used for research, evaluation, planning and other purposes by various users in terms of analysis and forecasting of the health status of individuals.

Automated medical diagnostic approaches are mainly based on a Machine Learning (ML) algorithm. Subsequently, they are trained to learn decision characteristics of a physician for an explicit disease and then they can be used to support physician decision making to diagnose future patients of the same disease [1, 2].Inappropriately, there is no common model that can be adjusted for the diagnosis of all kinds of diseases [3].

A large number of features that can surpass the number of data themselves often characterizes the data used in ML. This problem known as "the curse of dimensionality" creates a challenge for various ML applications for decision support. This can increase the risk of taking into account correlated or redundant attributes which can lead to lower classification accuracy [4, 5, 6].

Therefore, the process of eliminating irrelevant features is a vital phase for designing decision support systems with high accuracy. Therefore, the key objective of this paper is to design a FS approach to reduce dimension of CAD dataset and to obtain higher accuracy classification rates. The GA wrapped BN consists of a two-step process: In the first step, we generate a subset of features, the dimension of data is reduced using the GA algorithm running in parallel with BN. Accordingly, after the new subset feature model is obtained, a BN classifier is used to measure feature model accuracy. A10-fold cross-validation strategy has been used for validating the obtained model. Then, the proposed approach is also compared with four additional ML algorithms: SVM, MLP and C4.5. Additionally, the proposed algorithm is compared with other FS Algorithms.

The rest of the paper is planned as follows. The next section introduces a literature survey for CAD problem. Section 3 describes CAD and it is followed by a global introduction to FS strategies. In Section 4, The ML methods are introduced used to evaluate the accuracy of the feature model obtained by GA wrapped BN algorithm and the proposed approach is then explained and presented. In Section 5, experimental results are discussed and the conclusion is presented in Sections 6.

## 2. RELATED WORKS

A Multiple studies have been proposed in the context of Fuzzy Logic Rule based for diagnosis of CAD. In [8] proposed a system for detecting the risk level for heart disease. This system consists of two phases, the first is the generation of fuzzy rules, and the second is the construction of a rule inference engine based on rules generated. Tsipouras et al. [9] have proposed a four stage system for decision support based on fuzzy rules: 1) construction of a decision tree from the data, 2) the extraction of rules from the tree, 3) the transformation rules from the rough form to fuzzy one and 4 ) and the model optimization. They obtained a classification accuracy of 65%. Another work in this context [10] have developed a fuzzy system. They extracted the rules using an extraction method based on Rough Set Theory [11]. The rules then are selected and fuzzified, after that, they weighted the rules using the information of support. The results showed that the system is able to have a better prediction percentage of CAD than cardiologists and angiography. Three expert cardiologists validate the system. Patil et al. [12] have proposed an intelligent system for predicting heart attacks; in order to make the data ready to be analyzed they integrate them into a data warehouse. Once the data is in the data warehouse, they built clusters using the K-means method to build groups of similar individuals. Therefore, with the help of the algorithm MAFIA, the frequent sequences appropriate for the preaching of heart attacks were extracted. With the use of frequent sequences as training set and the back propagation algorithm, the neural network was used. The results were satisfied in terms of prediction. In this study [13], the authors have used techniques of data mining to study the risk factors that contribute significantly to the prediction of coronary artery syndromes. They assumed that the class is the diagnosis - with dichotomous values indicating the presence or absence of disease. They applied the binary regression. The data were taken from two hospitals of Karachi, Pakistan. For better performance of the model, a data reduction technique such as principal component analysis ACP was applied. Fidele et al [14] used techniques of artificial intelligence as the basis for evaluating risk factors for CAD. A two-layer perceptron that uses the Levenberg-Marquardt algorithm and back propagation. They have shown the efficiency of their system by applying it to the Long Beach data set.

## 3. CORONARY ARTERY DISEASE

CAD includes a multitude of diseases related to the heart and circulatory system. Cardiovascular disorders are the most common coronary artery disease, which relate to the arteries of the heart, and include, among others, angina, heart failure, myocardial infarction (heart attack) and stroke brain (stroke) that occur when the brain receives inadequate blood supply.

Like all medical fields and to prevent cardiovascular disease, one of the possible solutions is to make people aware of their CAD risks in advance and take preventive measures accordingly. According to experts, early detection of CAD at the stage of angina can prevent the death if the proper medication is given by the following. This is where the importance of developing a system for the diagnosis of CAD to assist the physicians to prevent from such Pathology. Studies that have been made to study risk factors for the CAD [15, 16], other studies that try to analyze the 12-lead ECG [17, 18] and 18-lead ECG [15].

Patients were evaluated using 14 features. The data set is taken from the Data Mining Repository of the University of California, Irvine (UCI) [19]. To end with the system is tested using Cleveland data sets. Features such as Age, sex, chest pain type, resting blood pressure, serum cholesterol in mg/dl, fasting blood sugar, resting electrocardiographic results, maximum heart rate achieved, exercise induced angina, ST depression, slope of the peak exercise ST segment, number of major vessels, thal and the diagnosis of heart disease are presented.

## 4. FEATURE SELECTION APPROACH

FS is an active area of research and in development in various applications (indexing and retrieval of images, genomic analysis, document analysis...). A large number of algorithms have been proposed in the literature for unsupervised, supervised and semi-supervised FS. According to Dash et al., [20] a selection process of attributes is usually composed of four steps illustrated in Figure 1.

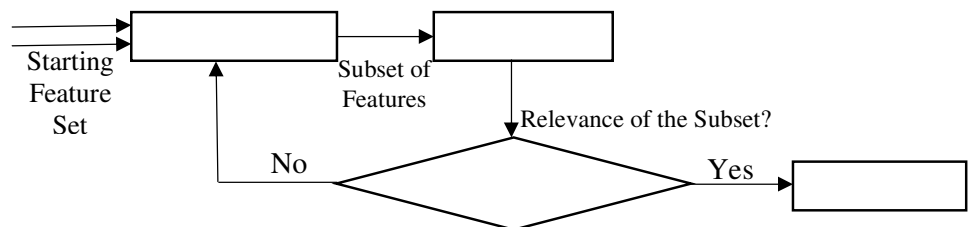

Figure 1.  Feature Selection process

The generation procedure allows, in each iteration, to generate a subset of attributes that will be evaluated in the second step of the selection procedure. This procedure of generation either can start with an empty set of attributes or with the set of all attributes or with a subset of attributes selected randomly. In the first two cases, attributes are added iteratively (forward selection) or removed (Backward selection) [21]. Some other algorithms hybrid both concepts as Sequential Forward Floating Selection technique that apply, after each step Forward, Backward steps while the selected subset improves the evaluation function [22]. In the third case, a new subset of attributes is created randomly at each iteration (random generation).Genetic algorithms, introduced by Holland in 1975 [23] are the most common methods used for random generation [24].

According to the evaluation criteria used in the selection process of attributes, we can distinguish between Wrapper approaches and Filter approaches. Wrapper Approaches use the classification accuracy rate as evaluation criteria [25]. Filter Approaches use an evaluation function based on the characteristics of the dataset, regardless of any classification algorithm, to select certain attributes or a subset of attributes (information measures, consistency measures, dependence measures and distance measures) [26][27].

## 4.1. The proposed Methodology

The random generation procedures explore randomly all $2^n$ subset candidates, which n is the number of features in the database. A subset is therefore not the result of an increase or decrease of features from the previous subset. This cannot stop the search when the evaluation function of a subset reached a local optimum. However, the $2^n$ subsets candidates are not all evaluated. Thus, a maximum number of iterations are imposed to ensure that computation time remains reasonable.

In our proposed Methodology, we make use of GA for the generation process Figure 2. GA is an adaptive heuristic search algorithm premised on the evolutionary ideas of natural selection and genetic. In our use of genetic algorithms, a subset of features is encoded as a chromosome. This group of chromosomes, i.e. population, is the search space of the algorithm. The fitness function used to evaluate performance of each chromosome (subset of features) to measure its nearness to solution. Accordingly, our study makes use of the BN Algorithm as a fitness function. The initial subsets of features (chromosomes) as parents are used to produce new subsets of features using genetic operators explicitly selection, mutation and crossover. At each iteration, the number of chromosomes in the population is always constant with removing features having lowest fitness value. This process is reiterated until a best subset of features is found or the maximum number of iterations is attained [28].We propose a new GA Wrapped BN methodology for identifying relevant features of CAD diagnosis Figure 2.

```
1 Split Dataset into 10 folds
2  For k=1 to 10
3     Test_DATA_all  =(1 fold for BN_Wrapper)
4     Train_data=(9 folds for training BN_evaluator)
5     for number_generation=1 to 40
6       Encode features as binary chromosomes
7       Generate randomly a population of 20 chromosomes
8       Evaluate accuracy of BN_evaluator generated in 7
9       Apply Binary Crossover with probability of 0.2
10      Apply Binary Mutation with probability of 0.09
11      Calculate the new Generated chromosomes Accuracy for BN_evaluator and
        compare it with 8
12      Replace chromosomes with lowest fitness of 7 by best chromosomes
        with highest fitness of 11
13     End for
14    Train BN_Wrapper with Train_data
15    Test BN_evaluator with test_Data_all
16    Calculate accuracy for k
17   end for
18 Calculate average accuracy for 10 folds
```

Figure 2. The proposed GA wrapped BN algorithm

In our model, we encoded features as binary strings of 1 and 0. In this pattern, 1 signifies selection of a feature and 0 means a non-selection. The number of genetic factor in the chromosome is equivalent to 14, which is the size of the features of CAD dataset. The population consist of the number of chromosomes in the search space and we have chosen 20 as the population size. Therefore, the first population (parent) is generated randomly. We make use of $BN_{evaluator}$ as a fitness function, generally a fitness function evaluate the relevance of chromosome. Accordingly, in our case the relevance of chromosomes is represented by classification accuracy of the $BN_{evaluator}$, so the aptness of a chromosome to be gathered depends on his accuracy rank obtained from the fitness function ($BN_{evaluator}$) at each iteration. For generation process of chromosomes, we make use of three genetic operators. The selection operator depends of the fitness function, so most relevant chromosomes are retained for each turn. Crossover operator make a two-point substring exchange between two chromosomes to generate two new chromosomes with probability of 0.2. In mutation Operator, the selected genetic factors are inverted to avoid the search process not to get fixed in local maxima. With a mutation probability of 0.09.

The proposed Algorithm make use of the three Operators, termination criteria for the selection process is the number of generations. It stopped when the number of generation is equal to 40 Figure 2. Line 5. The proposed algorithm have two loops imbricate in each other, the first loop Figure 2. Line 5 contains the generation process. First, the features are encoded as binary chromosomes Figure 2. Line6 then a new population is generated randomly Figure 2. Line 7. The generated population is evaluated using the fitness function ($BN_{evaluator}$). Formerly, we apply the Genetic Operators respectively Crossover Figure 2. Line 9 and mutation Figure 2. Line 10. The new generated population is evaluated with $BN_{evaluator}$ and then compared with that generated in Figure 2. Line 7. In the second, loop Figure 2. Line 2, the 9 folds set Figure 2. Line 4 with optimized new features of loop Figure 2. Line 5 is used as the train set for $BN_{evaluator}$. Having trained $BN_{wrapper}$ algorithm with this train set, the algorithm is used to classify instances of reserved test set Figure 2. Line 3. This process is reiterated 10 times with shifting folds recursively to achieve an average classification accuracy.

## 4.2. Feature Selection Techniques used in this study

In order to evaluate the efficacy of our proposed GN wrapped BN technique, we compare our methodology with another two FS wrapped methodologies. The two methodologies are based on BN algorithm. The first one uses the Best First Search (BFS) as a generation technique. BFS is a search algorithm that explores a graph by expanding the most promising node with the best score which will be evaluated using the wrapped BN [29].In the second methodology we generate the subset of features using the Sequential Floating Forward Search (SFFS). This technique is derived from the sequential forward generation techniques. The principle of such techniques is to add one or more attributes progressively. However, as they do not explore all possible subsets of attributes and cannot backtrack during the search, so they are suboptimal. SFFS after each step Forward, it applies Backward steps while the subset corresponding improves the efficacy of wrapped BN [30].

## 4.3. Machine Learning techniques used in our study

Our proposed methodology selects the most pertinent features from CAD dataset and it produces promising diagnosis accuracy. Nevertheless, to study the effectiveness of the selected features with other ML algorithms. In this section, we introduce generally those algorithms.

### 4.3.1. Naïve Bayes

One of the Bayesian approaches is NB. All of Bayesian approaches use Bayes formula (1). The main hypothesis of this kind of methods is independency of features. Thus, when features are dependent on each other, this algorithm produce a low classification accuracy [31].

$$P(A/B) = \frac{(P(B/A) * P(A))}{P(B)} \tag{1}$$

### 4.3.2. C4.5 Tree Algorithm

One of the important decision tree algorithms is C4.5 [32]. This algorithm can deal with all kinds of data. It uses pruning techniques to increase accuracy and Gain Ratio for selecting features. For instance, C4.5 can use a pruning algorithm such as reduce error pruning and it increases accuracy of the algorithm. One of its parameters is M, which the minimum number of instances that a leaf should have. The second one is C, it is threshold for confidence, and it is used for pruning process.

### 4.3.3. Support Vector Machine

SVM method is a supervised ML method, used for classification. It is widely used to produce a predicting model. For each given test input, SVM predicts which of two possible classes forms the input, making it a non-probabilistic binary linear classifier [33]. Given a training se of instance label pairs $(x_i, y_i)$, i=1,……,r, where $x_i \in R^n$ and $y \in [1, -1]^r$. SVM involves the resolution of the problem given by (2)

$$min_{w,b,\varepsilon} \frac{1}{2} w^T w + C \sum_{i=1}^{r} \varepsilon_i$$

With $y_i(w^T \varphi(x_i) + b) \geq 1 - \varepsilon_i, \varepsilon_i \geq 0$ (2)

In (2), $x_i$ are training vectors and they are mapped into a higher dimensional space by the function $\varphi$. The C is the decision parameter for the error term. SVM finds a linear separating hyper plane with the highest margin in the dimensional space. Accordingly, the solution of (2) permit only a linear separation solution. Conversely, the use of a kernel allows nonlinear separation using a kernel function (linear, polynomial, radial basis and sigmoid kernel). In our study, we use the polynomial kernel.

### 4.3.4. Multi-Layer Perceptron

MLP is feed-forward neural networks trained with the standard back-propagation algorithm [34]. It is supervised networks, so it learns based on an input data to conclude a desired response, so they are widely used for pattern classification. MLP contain one or two hidden layers. Obviously, the structure of MLP consists of an input and an output layer with one or more hidden layers. And for each node in one layer is linked to every node in the following layer with a weight of $w_{ij}$. In MLP, the learning task occurs while the weights of nodes are updated with the use of back propagation method and the amount of error in the output compared to the desired result. More explicitly, error in output node $j$ in the $n$th data point is represented by (3).

$$e_j(n) = d_j(n) - y_j(n) \tag{3}$$

In the equation, $d$ is the desired value and $y$ is the value produced by the MLP. Then, updating the weights (5) of the nodes based on those corrections, which minimize the entire output error, must be assessed and this is given by (4).

$$\varepsilon(n) = \frac{1}{2} \sum_j e_j^2(n) \tag{4}$$

$$\Delta w_{ji}(n) = -\delta \frac{\varepsilon(n)}{\partial v_j(n)} y_i(n) \tag{5}$$

Where $y_i$ the output of the previous neuron and $\vartheta$ is the learning rate. In our experiments, we uses a learning rate of 0.3 and training time as 500 iterations.

## 5. EXPERIMENTS AND RESULTS

For experimentation, we have chosen UCI CAD dataset, which is broadly accepted databases acquired from the UCI machine learning repository. In the testing phase, the testing dataset is given to the system to find the risk forecast of heart patients and achieved results are evaluated with the evaluation metric accuracy [35].

Accuracy is the typically used measure to evaluate the efficacy ML method; it is used to reckon how the test was worthy and consistent. In order to calculate these metric, we first compute some of the terms like, True positive, True negative, False negative and False positive based on Table 1., where TP is the True positive, TN is the True negative, FN is the False negative and FP is the False positive.

Table 1.  Confusion Matrix.

| Result of the diagnostic test | | Physician diagnosis | |
|---|---|---|---|
| | | Positive | Negative |
| Classifier Result | Positive | TP | FP |
| | Negative | FN | TN |

$$Accuracy = (TN + TP)/(TN + TP + FN + FP)$$

The experiments are approved and for each of the proposed wrappers. We first calculated average accuracies of the wrapper algorithms coupled with BN classifier. In addition, each wrapper strategy produces feature subsets and we made experiments to measure effectiveness of these features with the use of the described ML methods (Section 4.3) used for CAD diagnosis. The results of the experiments are evaluated with average Accuracy. In order to show the relevance of FS strategies, we involved the performance of ML classifiers without FS in Table 3. Moreover, Table 2 delivers the list of features for CAD dataset and the selected subset of features generated by the wrapper algorithms.

Table 2.  The selected features of CAD dataset by different wrapper techniques.

| N° | Feature | Selected Features by FS approaches | | | | | |
|---|---|---|---|---|---|---|---|
| | | GA wrapped BN | GA wrapped SVM | GA wrapped MLP | GA wrapped C4.5 | BFS wrapped BN | SFFS wrapped BN |
| 1 | Chest pain type: cp | ✓ | ✓ | ✓ | ✓ | | ✓ |
| 2 | Age | | ✓ | ✓ | | | |
| 3 | Sex | ✓ | | ✓ | | | |
| 4 | Resting blood pressure:  restbps | | | ✓ | | | |
| 5 | Cholesterol: chol | | | | | ✓ | |
| 6 | Fasting blood sugar: fbs | | | | ✓ | ✓ | |
| 7 | Resting electrocardiographic results: restescg | ✓ | | | | | ✓ |
| 8 | Maximum heart rate achieved: thalach | | | | | ✓ | ✓ |

| 9 | Exercise induced angina: exang | | ✓ | | | ✓ | |
|---|---|---|---|---|---|---|---|
| 10 | ST depression induced by exercise relative to rest: oldpeak | ✓ | ✓ | | | | ✓ |
| 11 | The slope of the peak exercise ST segment: slope | ✓ | ✓ | ✓ | | | |
| 12 | Number of major vessels(0-3) colored by fluoroscopy: ca | ✓ | ✓ | ✓ | ✓ | ✓ | ✓ |
| 13 | Thalium: thal | ✓ | ✓ | ✓ | ✓ | ✓ | ✓ |

Table 3. show results of average accuracies acquiredusing10-fold cross validation. It is recognizable from Table 3. that GA wrapper produces the most efficient feature model and therefore BN products a major diagnosis accuracy of 85.50%. Instead, it can be examined from Table 3. that feature selection algorithms produces a strength features compared to full dataset (without FS). Table 3. In addition, presents that other ML algorithms have satisfactory results. These classification performance results show that the feature model engendered with GA wrapper approach is powerful. Generally, results of our experiments and that from literature are presented in Figure 3. It is perceived that the proposed algorithm has the highest classification accuracy in the literature.

Table 3.  The performance evaluations of wrapper based feature selection algorithms.

| Wrapper Algorithms | Accuracy of different ML methods | | | |
|---|---|---|---|---|
| | BN | SVM | MLP | C4.5 |
| GA wrapper | 85.50 | 83.82 | 79.86 | 78.54 |
| BFS wrapper | 83.50 | 80.53 | 80.53 | 78.55 |
| SFFS wrapper | 84.49 | 83.17 | 77.89 | 78.22 |
| Without FS | 82.50 | 83.17 | 79.20 | 76.57 |

## 5.1. Discussion

Generally, results of our experiments and that from literature are presented in Figure 3. It is perceived that the proposed algorithm has the highest classification accuracy in the literature.

This study presents a GA wrapped BN classification model for diagnosis of CAD. The experimental results clearly show the efficiency of feature model selected by GA wrapper. In order to prove that, the proposed GA wrapper BN was compared with some other clinical systems from the literature. The algorithm was compared also with two wrapper FS methods (BFS and SFFS). The classification accuracies demonstrate the effectiveness of the features subset selected by GA.

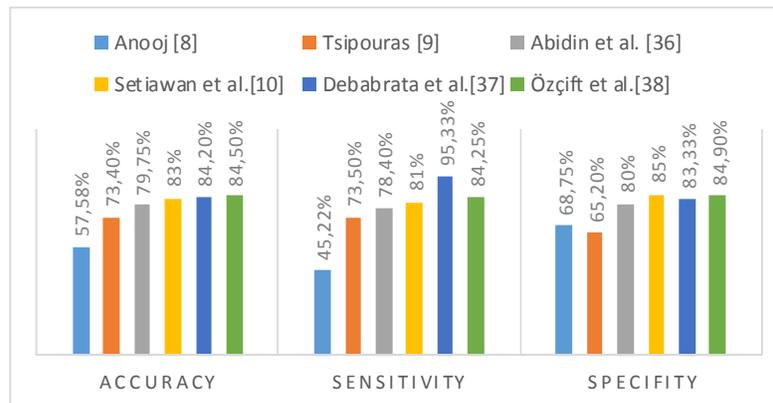

Figure 3.  Analysis of different Systems

## 6. Conclusions

We have presented a new GA wrapped BN FS algorithm for the diagnosis of the heart disease. The automatic process to generate the subset of features is an advantage of the proposed algorithm. The proposed algorithm for CAD patients contains two steps such as: (1) generation of a subset of features and (2) and the evaluation of the system using BN ML technique. The experimental results demonstrate the strength of the proposed GA wrapped BN algorithm for selecting the most relevant features for efficient diagnosis of CAD diseases. Generally automated disease diagnosis problems need a reduction of Features space step to achieve high accuracy performance. Consequently, the proposed algorithm is applied to CAD disease.

## Authors

Sidahmed MOKEDDEM was born in Mostaganem, Algeria, in 1989. He received the Master degree in computer engineering and informatics from the University of Oran, Oran, Algeria, in 2010. He is currently working toward the Ph.D. degree in computer science at the Department of computer science, University of Oran.

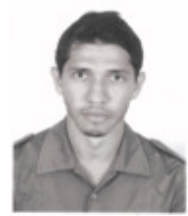

He is with the Laboratory of Oran LIO, Department of Computer Science, University of Oran. His current research interests include medical data mining, decision support systems in healthcare, and biomedical applications and bio-informatiques.

Baghdad ATMANI received his Ph.D. degree in computer science from the University of Oran (Algeria), in 2007. His interest field is Data Mining and Machine Learning Tools. His research is based on Knowledge Representation, Knowledge-based Systems and CBR, Data and Information Integration and Modelling, Data Mining Algorithms, Expert Systems and Decision Support Systems.

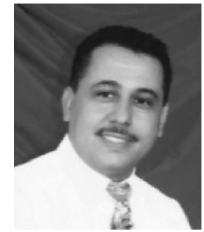

His research are guided and evaluated through various applications in the field of control systems, scheduling, production, maintenance, information retrieval, simulation, data integration and spatial data mining.

Mostefa MOKADDEM received the Engineer Diploma (1985), and the Magister (2008) in Computer Sciences at the University of Oran, Algeria. His is Assistant Professor at the Computer Science Department of University of Oran.

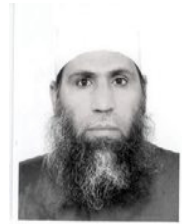

His current research interests are related with modeling methodologies, parallel/distributed simulation, Service Oriented Architecture, Knowledge Representation, Knowledge-based Systems, Data and Information Integration, Expert Systems and Decision Support Systems. His researches are guided and evaluated through various applications particularly in epidemic modeling and spatial data mining.